\documentclass[10pt, conference, compsocconf]{IEEEtran2}
\usepackage{graphicx}
\usepackage{color}
\usepackage[sort&compress,square,numbers]{natbib}
\usepackage{multirow} 
\usepackage{amsmath}
\usepackage{booktabs}
\usepackage{hyperref}
\usepackage{csquotes}
\usepackage{flushend}

\usepackage{amsmath}
\usepackage{hyperref}

\renewcommand{\paragraph}[1]{{\vspace{5pt}\noindent\bf #1}}
\renewcommand{\vec}[1]{\boldsymbol{#1}}

\begin{document}
\title{Audio Adversarial Examples: Targeted Attacks on Speech-to-Text}

\author{Nicholas Carlini \qquad David Wagner \\ University of California, Berkeley}
\maketitle

\begin{abstract}
We construct
targeted audio adversarial examples on automatic speech recognition.
Given any audio waveform,
we can produce another that is over 99.9\% similar,
but transcribes as any phrase we choose
(recognizing up to 50
characters per second of audio).
We apply our white-box iterative optimization-based attack to
Mozilla's implementation DeepSpeech end-to-end, and show it has a
100\% success rate.
The feasibility of this attack introduce a new domain to
study adversarial examples.
\end{abstract}

\section{Introduction}
As the use of neural networks continues to grow, it is
critical to examine their behavior in adversarial settings.
Prior work \cite{biggio2013evasion} has shown that neural
networks are vulnerable to
\emph{adversarial examples} \cite{szegedy2013intriguing},
instances $x'$ similar to a natural instance $x$, but classified
by a neural network
as any (incorrect) target $t$ chosen by the adversary.

Existing work on adversarial examples has focused
largely on the space of
images, be it image classification \cite{szegedy2013intriguing},
generative models on images \cite{kos2017adversarial},
image segmentation \cite{arnab2017robustness},
face detection \cite{sharif2016accessorize},
or reinforcement learning by manipulating the images the RL agent
sees \cite{behzadan2017vulnerability,huang2017adversarial}.
In the discrete domain, there has been some study of adversarial
examples over text classification \cite{jia2017adversarial} and
malware classification \cite{grosse2016adversarial,hu2017generating}.

There has been comparatively little study on the space of audio,
where the most common use is performing automatic speech recognition.
In automatic speech recognition, a neural network is
given an audio waveform $x$ and
perform the speech-to-text transform that gives the transcription $y$
of the phrase being spoken (as used in, e.g., Apple Siri,
Google Now, and Amazon Echo).

Constructing targeted adversarial examples on speech recognition
has proven difficult.
Hidden and inaudible voice commands \cite{carlini2016hidden,zhang2017dolphinatack,song2017inaudible} are
targeted attacks, but require synthesizing new audio and can not
modify existing audio (analogous
to the observation that neural networks can make high confidence predictions
for unrecognizable images \cite{nguyen2015deep}).
Other work has constructed standard untargeted adversarial examples on
different audio systems \cite{kereliuk2015deep,gong2017crafting}.
The current state-of-the-art targeted attack on automatic speech
recognition is
Houdini \cite{cisse2017houdini}, which can only construct
audio adversarial examples targeting phonetically similar
phrases, leading the authors to state
\begin{displayquote}
  {
    targeted attacks seem to be much
    more challenging when dealing with speech recognition systems
    than when we consider artificial
    visual systems.}
\end{displayquote}

\begin{figure}
  \centering
  \includegraphics[scale=.34]{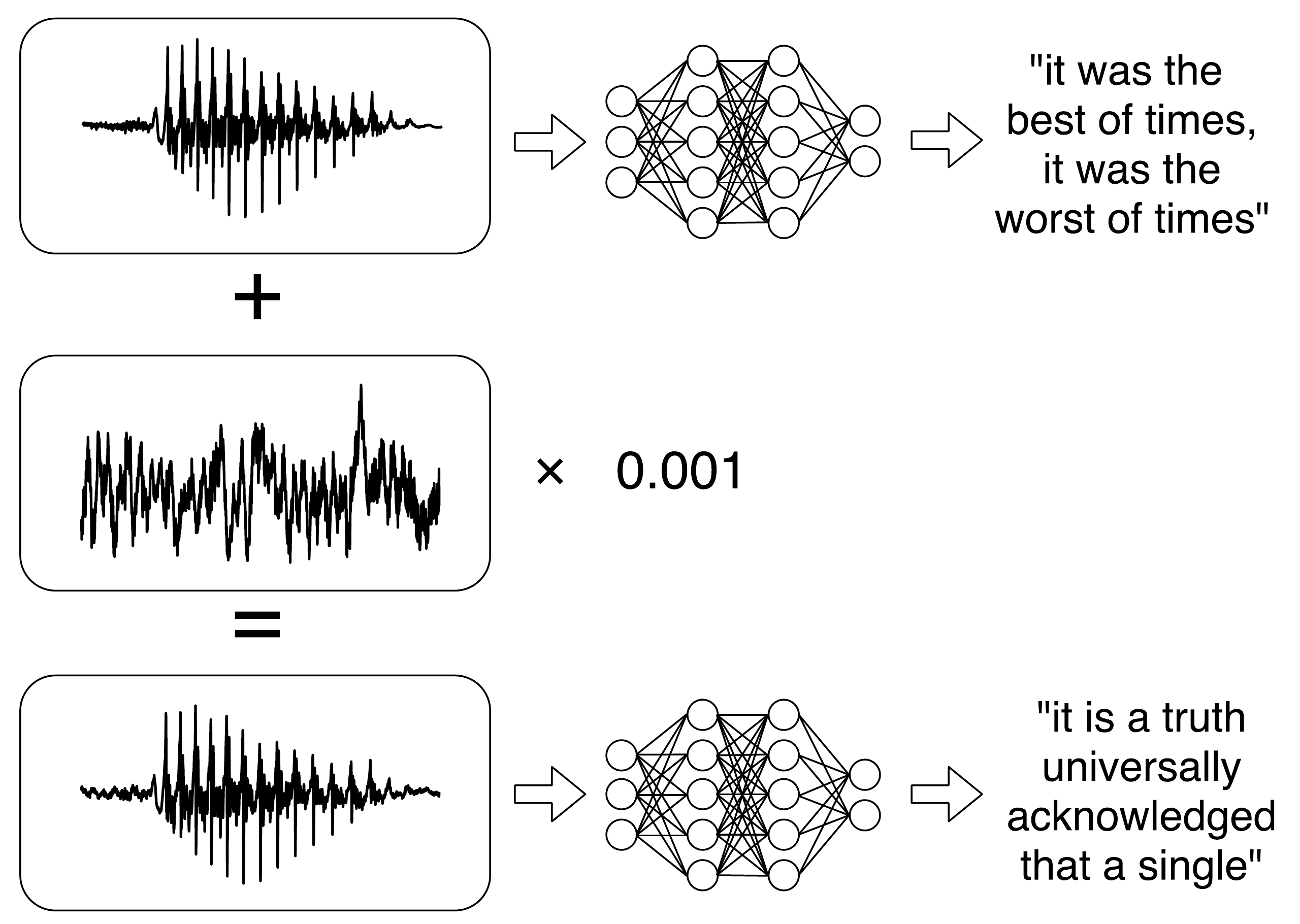}
  \caption{Illustration of our attack: given any waveform,
    adding a small perturbation 
    makes the result
    transcribe as any desired target phrase.}
  \label{fig:intro}
\end{figure}

\paragraph{Contributions.}
In this paper, we demonstrate that targeted
adversarial examples exist in the audio domain by attacking
DeepSpeech \cite{hannun2014deep}, a state-of-the-art speech-to-text
transcription neural network.
Figure~\ref{fig:intro} illustrates our attack: given any natural
waveform $x$, we are able to construct a perturbation $\delta$ that
is nearly inaudible but so that $x + \delta$ is recognized
as \textbf{any} desired phrase. We are able to achieve this by
making use of strong, iterative, optimization-based attacks based
on the work of
\cite{carlini2017towards}.

Our white-box attack is end-to-end, and operates
directly on the raw samples that are used as input to the classifier.
This requires optimizing through the MFC pre-processing
transformation, which is
has been proven to be difficult \cite{carlini2016hidden}.
Our attack works with $100\%$ success, regardless of the
desired transcription or initial source audio sample.

By starting with an arbitrary waveform, such as
music, we can embed speech into audio that should not be recognized
as speech; and by choosing
silence as the target, we can hide audio from a speech-to-text
system.

Audio adversarial examples give a new domain to explore these
intriguing properties of neural networks.
We hope others will build
on our attacks to further study this field.
To facilitate future work,
we make our code and dataset available\footnote{\url{http://nicholas.carlini.com/code/audio_adversarial_examples}\label{foot}}.
Additionally, we encourage the reader
to listen to our audio adversarial examples.

\section{Background}

\paragraph{Neural Networks \& Speech Recognition.}
A neural network is a differentiable
parameterized function $f(\cdot)$. Its parameters can be
updated by gradient descent to learn any function.

We represent audio as a $N$-dimensional vector $\vec{x}$. Each
element $x_i$ is a signed 16-bit value, sampled at 16KHz. To reduce
the input dimensionality, the Mel-Frequency Cepstrum (MFC) transform
is often used as a preprocessing step \cite{hannun2014deep}.
The MFC splits the waveform into
50 \emph{frames} per second, and maps each frame to the frequency domain.

Standard classification neural networks take one input and produce an
output probability distribution over all output labels.
However, in the case of speech-to-text systems, there are exponentially
many possible labels, making it computationally infeasible to
enumerate all possible phrases.

Therefore, speech recognition systems often use 
Recurrent Neural Networks (RNNs)
to map an audio waveform to a sequence
of probability distributions over individual characters,
instead of over complete phrases.
An RNN is a function which maintains a state vector $s$ with $s_0 = \vec{0}$
and $(s_{i+1}, y^{i}) = f(s_{i}, x_i)$,
where the input $x_i$ is one frame of
input, and each output $y^i$ is a probability distribution over which character was
being spoken during that frame.

We use the DeepSpeech \cite{hannun2014deep} speech-to-text system (specifically,
Mozilla's implementation  \cite{mozilla2017deepspeech}).
Internally, it consists of a preprocessing layer which computes the
MFC followed by a recurrent neural network
using LSTMs \cite{hochreiter1997long}.

\paragraph{Connectionist Temporal Classiﬁcation}
(CTC) \cite{graves2006connectionist} is a method of training a sequence-to-sequence neural network
when the alignment
between the input and output sequences is not known. DeepSpeech uses CTC
because the inputs are
an audio sample of a person speaking, and the unaligned transcribed sentences,
where the exact position of each word in the audio sample is not known.

We briefly summarize the key details and notation.
We refer readers to \cite{hannun2017sequence} for an excellent survey
of CTC.

Let $\mathcal{X}$ be the input domain --- a single frame of input ---
and $\mathcal{Y}$ be the range --- the characters a-z, space,
and the special $\epsilon$ token (described below).
Our neural network $f~:~\mathcal{X}^N \to [0,1]^{N \cdot |\mathcal{Y}|}$ takes a sequence of $N$
frames $x \in \mathcal{X}$ and returns a probability distribution
over the output domain for each frame.
We write $f(\vec{x})^i_j$ to mean that the probability of
frame $x_i \in \mathcal{X}$
having label $j \in \mathcal{Y}$. We use $\vec{p}$ to denote a
phrase: a sequence of characters $\langle{}p_i\rangle{}$, where
each $p_i \in \mathcal{Y}$.

While $f(\cdot)$ maps every frame to a probability distribution over the characters,
this does not directly give a probability distribution over all
\emph{phrases}. The probability of a phrase is defined as a function
of the probability of each character.

We begin with two short definitions.
We say that a sequence $\pi$ \emph{reduces to}
$\vec{p}$ if starting with $\pi$ and making the following
two operations (in order) yields $\vec{p}$:
\begin{enumerate}
\item Remove all sequentially duplicated tokens.
\item Remove all $\epsilon$ tokens.
\end{enumerate}
For example, the sequence $a\; a\; b\; \epsilon\; \epsilon\; b$ reduces to $a\; b\; b$.

Further, we say that $\pi$ is an alignment of $\vec{p}$ 
with respect to $\vec{y}$ (formally: $\pi \in \Pi(\vec{p},\vec{y})$) if
(a) $\pi$ \emph{reduces to} $\vec{p}$, and (b) the length
of $\pi$ is equal to the length of $\vec{y}$.
The probability of alignment $\pi$ under $\vec{y}$ is the product of the
likelihoods of each of its elements:
\[ \Pr(\pi | \vec{y}) = \prod\limits_i \vec{y}^i_{\pi^i} \]

With these definitions, we can now define the probability of a given phrase
$\vec{p}$ under the distribution $\vec{y} = f(\vec{x})$ as
\[\Pr(\vec{p} | \vec{y}) = \sum\limits_{\pi \in \Pi(\vec{p},\vec{y})} \Pr(\pi | \vec{y})
= \sum\limits_{\pi \in \Pi(\vec{p},\vec{y})} \prod\limits_i \vec{y}^i_{\pi^i}\]

As is usually done, the loss function used to train the network is the
negative log likelihood of the desired phrase:
\[\text{CTC-Loss}(f(\vec{x}),\vec{p}) = - \log \Pr(\vec{p} | f(\vec{x})).\]
Despite the exponential search space, this loss can be computed
efficiently with dynamic programming \cite{graves2006connectionist}.

Finally, to \emph{decode} a vector $\vec{y}$ to a phrase $\vec{p}$, we
search for the phrase $\vec{p}$ that best aligns to $\vec{y}$. 
\[C(\vec{x}) = \mathop\text{arg max}\limits_{\vec{p}} \Pr(\vec{p} | f(\vec{x})). \]

Because computing $C(\cdot)$ requires searching an exponential space, it is typically
approximated in one of two ways.
\begin{itemize}
\item
\emph{Greedy Decoding} searches for the most likely alignment (which is easy
 to find) and then reduces this alignment to obtain the transcribed phrase:
\[C_\text{greedy}(\vec{x}) = \text{reduce}(\mathop\text{arg max}\limits_\pi \Pr(\pi | f(\vec{x}))) \]

\item
  \emph{Beam Search Decoding} simultaneously evaluates the likelihood of multiple
  alignments $\pi$ and then chooses the most likely phrase $\vec{p}$ under these
  alignments.  We refer
  the reader to \cite{graves2006connectionist} for a complete algorithm description.
\end{itemize}

\paragraph{Adversarial Examples.}
Evasion attacks have long been studied on machine learning classifiers
\cite{lowd2005adversarial,barreno2006can,barreno2010security}, and
are practical against many types of
models \cite{biggio2013evasion}.

When discussion neural networks, these evasion attacks are referred to
as \emph{adversarial examples} \cite{szegedy2013intriguing}: for any
input $x$, it is possible to construct a sample
$x'$ that is similar
to $x$ (according to some metric) but so that $C(x) \ne C(x')$ \cite{biggio2013evasion}.
In the audio domain, these untargeted adversarial example are
usually not interesting:
causing a speech-to-text system to transcribe ``test sentence'' 
as the misspelled ``test \emph{sentense}'' does little to help an adversary.

\paragraph{Targeted Adversarial Examples} are a more powerful attack:
not only must the classification of $x$ and $x'$ differ, but
the network must assign a specific label (chosen by the adversary)
to the instance $x'$.
The purpose of this paper is to show that
targeted adversarial examples are possible
with only slight distortion on speech-to-text systems.

\section{Audio Adversarial Examples}

\subsection{Threat Model \& Evaluation Benchmark}

\paragraph{Threat Model.} Given an audio waveform $x$, and target
transcription $y$, our task is
to construct another audio waveform $x' = x+\delta$ so that $x$ and $x'$ sound
similar (formalized below), but so that $C(x') = y$.
We report success only if the output of the network matches exactly
the target phrase (i.e., contains no misspellings or extra characters).

We assume a white-box setting where the adversary has complete knowledge
of the model and its parameters. This is the threat model taken in most
prior work \cite{goodfellow2014explaining}. Just as
later work in the space of images showed black-box attacks are possible
\cite{papernot2016practical,ilyas2017query};
we expect that our attacks can be extended to black-box attacks.
Additionally, we assume our adversarial examples are directly
classified without any noise introduced (e.g., by playing them over-the-air
and then recording them with a microphone). Initial work on image-based
adversarial examples also made this same assumption, which was later
shown unnecessary \cite{kurakin2016adversarial,athalye2017synthesizing}.


\paragraph{Distortion Metric.}
How should we quantify the distortion introduced by a perturbation
$\delta$? In the space of images, despite some debate
\cite{rozsa2016adversarial}, most of the community has settled on
$l_p$ metrics
\cite{carlini2017towards}, most often using $l_\infty$
\cite{goodfellow2014explaining,madry2017towards}, the maximum amount any pixel
has been changed. We follow this convention for our audio attacks.

We measure distortion in Decibels (dB): a logarithmic scale that measures
the relative loudness of an audio sample:
\[dB(x) = \max_i \; 20 \cdot \log_{10}(x_i).\]
To say that some signal is ``10 dB'' is only meaningful when comparing it
relative to some other reference point. In this paper, we compare the dB level
of the distortion $\delta$ to the
original waveform $x$. To make this explicit, we write
\[dB_x(\delta) = dB(\delta) - dB(x).\]

Because the perturbation introduced is \emph{quieter} than
the original signal, the distortion is a negative number, where
smaller values indicate quieter distortions. 

While this metric may not be a perfect measure of distortion,
as long as the perturbation is small enough, it will be imperceptible
to humans. We encourage the reader to listen to our adversarial
examples to hear how similar they sound. Alternatively, later, in Figure~\ref{fig:twowave},
we visualize two waveforms which transcribe to different phrases overlaid.

\paragraph{Evaluation Benchmark.}
To evaluate the effectiveness of our attack, we construct
targeted audio adversarial examples on the first $100$ test instances
of the Mozilla Common Voice dataset. For each sample, we target $10$
different \emph{incorrect} transcriptions, chosen at random
such that (a) the transcription is incorrect, and (b) it is theoretically
possible to reach that target.

\subsection{An Initial Formulation}
\label{sec:attack}
As is commonly done \cite{biggio2013evasion,szegedy2013intriguing},
we formulate the problem of constructing an
adversarial example as an optimization problem: given a
natural example $x$ and any target phrase $t$, we
solve the formulation
\begin{align*}
  \text{minimize} \;& dB_x(\delta)\\
  \text{such that} \;& C(x+\delta) = t \\
  & x+\delta \in [-M,M]
\end{align*}
Here M represents the maximum representable value ($2^{15}$ in our case).
This constraint can be handled by clipping the values of $\delta$;
for notational simplicity we omit it from future formulation.
Due to the non-linearity of the constraint $C(x+\delta)=t$,
standard gradient-descent techniques do not work well with this
formulation. 

Prior work \cite{szegedy2013intriguing}
has resolved this through the reformulation
\begin{align*}
  \text{minimize} \;& dB_x(\delta) + c \cdot \ell(x+\delta, t)
\end{align*}
where the loss function $\ell(\cdot)$ is constructed so that
$\ell(x', t) \le 0 \iff C(x') = t.$
The parameter $c$
trades off the relative importance of being adversarial and
remaining close to the original example.

Constructing a loss function $\ell(\cdot)$ with this property
is much simpler in the domain of images than in the domain of
audio; on images, $f(x')_y$ directly corresponds to the probability
of the input $x'$ having label $y$.
In contrast, for audio, we use a second decoding step to compute
$C(x')$, and so constructing a loss function is nontrivial.

To begin, we use the CTC loss as the loss function:
$\ell(x', t) = \text{CTC-Loss}(x', t).$
For this loss function, one direction of the implication
holds true (i.e.,
$\ell(x', t) \le 0 \implies C(x') = t$) but the converse
does not.
Fortunately, this means that the resulting solution will still be
adversarial, it just may not be minimally perturbed.

The second difficulty we must address is that when using a $l_\infty$
distortion metric, this
optimization process will often oscillate around a solution
without converging \cite{carlini2017towards}. Therefore, 
instead we initially solve the formulation
\begin{align*}
  \text{minimize} \;& |\delta|_2^2 + c \cdot \ell(x+\delta, t) \\
  \text{such that} \; & dB_x(\delta) \le \tau
\end{align*}
for some sufficiently large constant $\tau$.
Upon obtaining a partial solution $\delta^*$ to the above problem, we reduce $\tau$ and
resume minimization, repeating until no solution can be found.


To solve this formulation, we differentiate through the entire classifier
to generate our adversarial examples --- starting from
the audio sample, through the MFC, and neural network,
to the final loss.
We solve the minimization problem over the complete audio sample
simultaneously.
This is in contrast with prior work on hidden
voice commands \cite{carlini2016hidden}, which were
generated sequentially, one frame at a time.
We solve the minimization problem with the Adam \cite{kingma2014adam}
optimizer using a
learning rate of $10$, for a maximum of $5,000$ iterations.

\paragraph{Evaluation.}
We are able to generate targeted adversarial examples with $100\%$
success for each of the source-target pairs with a mean perturbation
of $-31$dB. For comparison, this is roughly the difference between
ambient noise in a quiet room and a person talking \cite{smith1997scientist}. We encourage
the reader to listen to our audio adversarial examples\textsuperscript{\ref{foot}}.
The $95\%$ interval for distortion ranged from $-15$dB to $-45$dB.

The longer a phrase is, the more difficult it is to target: every extra
character requires approximately a $0.1$dB increase in distortion. However,
conversely, we observe that the longer the initial source phrase is, the
\emph{easier} it is to make it target a given transcription. These two
effects roughly counteract each other (although we were not able to measure
this to a statistically significant degree of certainty).

Generating a single adversarial example requires approximately one hour
of compute time on commodity hardware (a single NVIDIA 1080Ti).
However, due to the massively parallel nature of GPUs, we
are able to construct $10$ adversarial examples simultaneously, reducing
the time for constructing any given adversarial example to only a few minutes.\footnote{Due to implementation difficulties, after constructing
  adversarial examples simultaneously, we must fine-tune them individually
  afterwards.}

\subsection{Improved Loss Function}
\label{sec:cwloss}

Carlini \& Wagner \cite{carlini2017towards} demonstrate that the
choice of loss function impacts the final distortion
of generated adversarial examples by a factor of $3$ or more.
We now show the same holds in the audio domain, but to a lesser extent.
While CTC loss is highly useful for training the neural network,
we show that a carefully designed loss function allows generating
better lower-distortion adversarial examples.
For the remainder of this section, we focus on generating adversarial examples
that are only effective when using greedy decoding.

In order to minimize the CTC loss (as done in \S~\ref{sec:attack}),
an optimizer will make \emph{every} aspect
of the transcribed phrase more similar to the target phrase. That is,
if the target phrase is ``ABCD'' and we are already decoding to ``ABCX'',
minimizing CTC loss will still cause the ``A'' to be more ``A''-like,
despite the fact that the only important change we require is for the
``X'' to be turned into a ``D''.

This effect of making items classified more strongly as the desired
label despite already having that label led to the design of
a more effective loss function:
\[\ell(y,t) = \max \left(y_t - \mathop{\max}\limits_{t' \ne t} y_{t'}, 0 \right).\]
Once the probability of item $y$ is larger than any other item,
the optimizer no longer sees a reduction in loss by making it more strongly
classified with that label.

We now adapt this loss function to the audio domain. Assume we were
given an alignment $\pi$ that aligns the phrase $\vec{p}$ with the
probabilities $\vec{y}$.
Then the loss of this sequence is
\[L(\vec{x}, \pi) = \sum\limits_i \ell(f(\vec{x})^i,\pi_i). \]

We make one further improvement on this loss function. The
constant $c$ used in the minimization formulation determines the relative
importance of being close to the original symbol versus being adversarial.
A larger value of $c$ allows the optimizer to place more emphasis
on reducing $\ell(\cdot)$.

In audio, consistent with prior work \cite{carlini2016hidden} we observe
that certain characters are more difficult for the transcription to
recognize. When we choose only one constant $c$ for the complete phrase,
it must be large enough so that we can make the most difficult character
be transcribed correctly. This forces $c$ to be larger
than necessary for the easier-to-target segments.
To resolve this issue, we instead use the following formulation:
\begin{align*}
  \text{minimize} \; & |\delta|_2^2 + \sum\limits_i c_i \cdot L_i(x+\delta, \pi_i) \\
  \text{such that} \; & dB_x(\delta) < \tau
\end{align*}
where $L_i(\vec{x},\pi_i) = \ell(f(\vec{x})^i,\pi_i)$.
Computing the loss function requires choice of an alignment $\pi$.
If we were not concerned about runtime efficiency, in principle we could
try all alignments $\pi \in \Pi(\vec{p})$ and select the best one.
However, this is computationally prohibitive.

Instead, we use a two-step attack:
\begin{enumerate}
\item First, we let $x_0$ be an adversarial example found using the CTC loss
(following \S\ref{sec:attack}).
CTC loss explicitly constructs an alignment during
decoding.
We extract the alignment $\pi$ that is induced by $x_0$
(by computing $\pi = \text{arg max}_i\,\, f(x_0)^i$).
We fix this alignment $\pi$ and use it as the target in the second step.

\item Next, holding the alignment $\pi$ fixed, we generate a less-distorted
  adversarial example $x'$ targeting the alignment $\pi$ using
  the improved loss function
above to minimize
$|\delta|_2^2 + \sum_i c_i \cdot \ell_i(x+\delta, \pi)$, starting
gradient descent at the initial point $\delta = x_0-x$.
\end{enumerate}

\paragraph{Evaluation.}
We repeat the evaluation from Section~\ref{sec:attack} (above), and
generate targeted adversarial examples for the first 100 instances of
the Common Voice test set. We are able to reduce the mean distortion
from $-31$dB to $-38$dB. However, the adversarial examples we generate
are now only guaranteed to be effective against a greedy decoder;
against a beam-search decoder, the transcribed phrases are often more
similar to the target phrase than the original phrase, but do not
perfectly match the target.

\begin{figure}
  \includegraphics[clip, trim=1in 0cm 0.5cm 0cm, scale=.66]{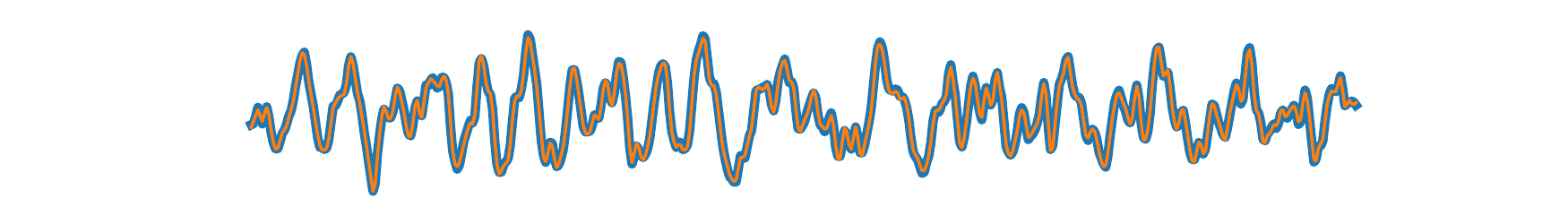}
  \caption{Original waveform (blue, thick line) with adversarial waveform (orange, thin line)
    overlaid; it is nearly impossible to notice a difference.
    The audio waveform was chosen randomly from the
    attacks generated and is 500 samples long. }
  \label{fig:twowave}
\end{figure}

Figure~\ref{fig:twowave} shows two waveforms overlaid; the blue, thick line
is the original waveform, and the orange, thin line the modified adversarial
waveform. This sample was chosen randomly from among the training
data, and corresponds to a distortion of $-30$dB. Even visually, these
two waveforms are nearly indistinguishable.

\subsection{Audio Information Density}
Recall that the input waveform is converted into 50 \emph{frames} per
second of audio, and DeepSpeech outputs one probability distribution of
characters per frame. This places the theoretical maximum density
of audio at 50 characters per second.
We are able to generate adversarial examples that produce output
at this maximum rate.
Thus, short audio clips can transcribe to a long textual phrase.

The loss function $\ell(\cdot)$ is simpler in this setting.
The \emph{only} alignment of $\vec{p}$ to $\vec{y}$ is the assignment
$\pi=\vec{p}$. This means that the logit-based loss function can be applied
directly without first heuristically finding an alignment; any other
alignment would require omitting some character.

We perform this attack and find it is effective, although it requires a
mean distortion of $-18$dB.

\subsection{Starting from Non-Speech}

Not only are we able to construct adversarial examples that cause
DeepSpeech to transcribe the incorrect text for a person's speech,
we are also able to begin with arbitrary non-speech audio sample
and make that recognize as any target phrase.
No technical novelty on top of what was developed above is required
to mount this attack: we only let the initial audio waveform be
non-speech.

To evaluate the effectiveness of this attack, we take five-second
clips from classical music (which contain no speech) and target
phrases contained in the Common Voice dataset. We have found that
this attack requires more computational effort (we perform
$20,000$ iterations of gradient descent) and the total distortion
is slightly larger, with a mean of $-20$dB.

\subsection{Targeting Silence}

Finally, we find it is possible to \emph{hide} speech by adding adversarial
noise that causes DeepSpeech to transcribe nothing. While performing
this attack without modification (by just targeting the empty phrase)
is effective, we can slightly improve on this if we define silence to
be an arbitrary length sequence of only the space character repeated. With
this definition, to obtain silence, we should let
\[\ell(\vec{x}) = \sum\limits_i \max\left(
\mathop{\max}\limits_{t \in\{\epsilon,`` "\}} f(\vec{x})^i_{t} -
\mathop{\max}\limits_{t' \not\in\{\epsilon,`` "\}} f(\vec{x})^i_{t'}, 0\right). \]

We find that targeting silence is easier than targeting
a specific phrase: with distortion less than $-45$dB below the original signal,
we can turn any phrase into silence.

This partially explains why it is easier to construct adversarial examples when
starting with longer
audio waveforms than shorter ones: because the longer phrase contains
more sounds, the adversary can silence the ones that are not
required and obtain a subsequence that nearly matches the target. In
contrast, for a shorter phrase, the adversary must synthesize new
characters that did not exist previously.

\section{Audio Adversarial Example Properties}

\subsection{Evaluating Single-Step Methods}

In contrast to prior work which views adversarial examples as ``blind spots''
of a neural network, Goodfellow \emph{et al.} \cite{goodfellow2014explaining}
argue that adversarial examples
are largely effective due to the locally linear nature of neural networks.

\begin{figure}
  \centering
  \includegraphics[scale=.75]{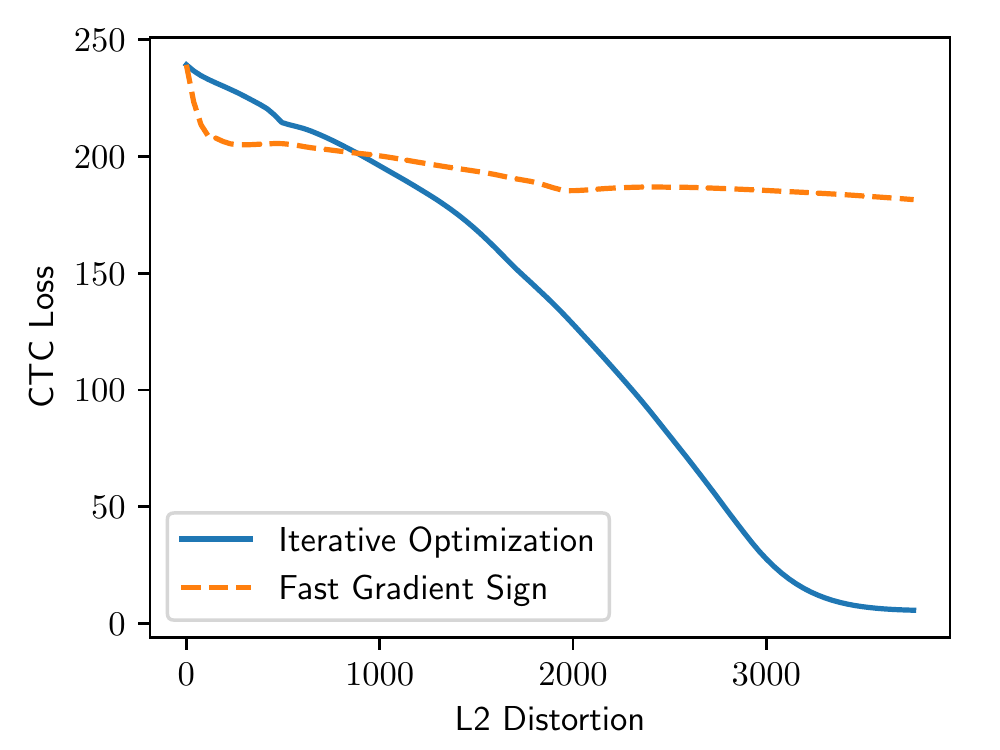}
  \caption{CTC loss when interpolating between the original audio sample
    and the adversarial example (blue, solid line), compared to traveling
    equally far in the direction suggested by the fast gradient sign
    method (orange, dashed line). Adversarial examples exist far enough away from the original
    audio sample that solely relying on the local linearity of neural networks is
    insufficient to construct targeted adversarial examples.}
  \label{fig:fgm}
\end{figure}

The Fast Gradient Sign Method (FGSM) \cite{goodfellow2014explaining}
demonstrates that this is true in the
space of images. FGSM
takes a single step in the direction of the gradient of the loss function.
That is, given network $F$ with loss function $\ell$, we compute the
adversarial example as
\[x' \gets x - \epsilon \cdot \text{sign}(\nabla_x \ell(x,y)).\]
Intuitively, for each pixel in an image, this attack asks ``in which direction should we
modify this pixel to minimize the loss?'' and then taking a small step
in that direction for every pixel simultaneously.
This attack can be applied directly to audio, changing individual samples
instead of pixels.

However, we find that this type of single-step attack is not effective on audio
adversarial examples: the inherent non-linearity introduced in computing the
MFCCs, along with the depth of many rounds of LSTMs, introduces a large degree of
non-linearity in the output.

In Figure~\ref{fig:fgm} we compare the value of the CTC loss when traveling
in the direction of a known adversarial example, compared to traveling in the fast
gradient sign direction. While initially (near the source audio sample),
the fast gradient direction is more effective at reducing
the loss function, it quickly plateaus and does not decrease
afterwards. On the other hand, using iterative optimization-based attacks
find a direction
that eventually leads to an adversarial example.
(Only when the CTC loss is below 10 does the phrase have the
correct transcription.)

We do, however, observe that the FGSM can be used to produce \emph{untargeted}
audio adversarial examples, that make a phrase misclassified (although
optimization methods again can do so with less distortion).

\subsection{Robustness of Adversarial Examples}

The minimally perturbed adversarial examples we construct in
Section~\ref{sec:attack} can be made non-adversarial by trivial
modifications to the input.
Here, we demonstrate here that it is possible to construct adversarial
examples robust to various forms of noise.

\paragraph{Robustness to pointwise noise.}
Given an adversarial example $x'$, adding pointwise random noise $\sigma$
to $x'$ and returning $C(x+\sigma)$ will cause $x'$ to lose its
adversarial label, even if the distortion $\sigma$ is small enough to
allow normal examples to retain their classification.

We generate a high confidence adversarial example
$x'$ \cite{biggio2013evasion,carlini2017towards}, and make use of
Expectation over Transforms \cite{athalye2017synthesizing} to generate
an adversarial example robust to this synthetic noise at $-30dB$.
The adversarial perturbation increases by approximately $10$dB when
we do this.

\paragraph{Robustness to MP3 compression.}
%
Following \cite{athalye2018obfuscated}, we make use of the straight-through
estimator \cite{bengio2013estimating} to construct adversarial examples
robust to MP3 compression.
We generate an adversarial example
$x'$ such that $C(\text{MP3}(x'))$ is classified as the target label by
computing gradients of the CTC-Loss assuming that the gradient of the MP3
compression is the identity function. While individual gradient steps are
likely not correct, in aggregate the gradients average out to become useful.
This allows us to generate adversarial examples with approximately $15dB$ larger
distortion that remain robust to MP3 compression.


\section{Open Questions}

\paragraph{Can these attacks be played over-the-air?}
Image-based adversarial examples have been shown to be feasible in the
physical world \cite{kurakin2016adversarial,athalye2017synthesizing}. In the
audio space, both hidden voice commands and Dolphin Attack's inaudible voice commands
are effective over-the-air when played by a speaker and recorded by
a microphone \cite{carlini2016hidden,zhang2017dolphinatack}.

The audio adversarial examples we construct in this paper
do not remain adversarial after being
played over-the-air, and therefore present a limited
real-world threat; however, just as the initial work on image-based
adversarial examples did not consider the physical channel and only
later was it shown to be possible, we believe further work will be able
to produce audio adversarial examples that are effective over-the-air.

\paragraph{Do universal adversarial perturbations \cite{moosavi2016universal} exist?}
One surprising observation is that on the space of images,
it is possible to construct a single perturbation $\delta$ that when
applied to an arbitrary image $x$ will make its classification incorrect.
These attacks would be powerful on audio, and would correspond
to a perturbation that could be played to cause any other waveform
to recognize as a target phrase.

\paragraph{Are audio adversarial examples transferable?}
That is,
given an audio sample $x$, can we generate a
single perturbation $\delta$ so that $f_i(x+\delta)=y$ for multiple
classifiers $f_i$?
Transferability is believed to be a fundamental property of neural
networks \cite{papernot2016transferability},
significantly complicates constructing robust defenses \cite{carlini2017magnet},
and allows attackers to mount black-box attacks \cite{liu2016delving}.
Evaluating transferability
on the audio domain is an important direction for future work.

\paragraph{Which existing defenses can be applied audio?}
To the best of our knowledge, all existing defenses to adversarial
examples
have only been evaluated on image domains. If the defender's
objective is to produce a robust neural network, then it should
improve resistance to adversarial examples on all domains,
not just on images. Audio adversarial examples give another
point of comparison.



\section{Conclusion}
We demonstrate targeted audio adversarial examples are effective
on automatic speech recognition. With
optimization-based attacks applied
end-to-end,
we are able to turn any audio
waveform into any target transcription with $100\%$ success
by only adding a slight distortion. We can cause audio to
transcribe up to 50 characters per second (the theoretical maximum),
cause music to transcribe as arbitrary speech, and hide speech
from being transcribed.

We present preliminary evidence that audio adversarial examples have
different properties from those on images by showing that
linearity does not hold on the audio domain.
We hope that future work will continue to
investigate audio adversarial examples, and
separate the fundamental properties of adversarial examples
from properties which occur only on image recognition.

\section*{Acknowledgements}
This work was supported by National Science
Foundation award CNS-1514457, Qualcomm,
and the Hewlett Foundation through the Center for Long-Term
Cybersecurity.



{\footnotesize
\bibliographystyle{abbrvnat}
\bibliography{paper}
}

\end{document}